\documentclass[10pt,twocolumn,letterpaper]{article}

\usepackage{iccv}              % To produce the CAMERA-READY version
\usepackage{footmisc}
\usepackage[accsupp]{axessibility}
\definecolor{iccvblue}{rgb}{0.21,0.49,0.74}
\usepackage[pagebackref,breaklinks,colorlinks,allcolors=iccvblue]{hyperref}
\usepackage{multicol, multirow}
\usepackage{amsmath}
\usepackage{verbatim}
\usepackage{romanbar}
\usepackage{textcomp}
\usepackage{tikz}

\def\paperID{7863} % *** Enter the Paper ID here
\def\confName{ICCV}
\def\confYear{2025}

\title{AdaptiveAE: An Adaptive Exposure Strategy for HDR Capturing \\ in Dynamic Scenes}

\author{
    Tianyi Xu$^{1,3,4*}$ \quad
    Fan Zhang$^1$ \quad
    Boxin Shi$^{3,4\dagger}$ \quad
    Tianfan Xue$^{2,1\dagger}$ \quad
    Yujin Wang$^{1\dagger}$
    \\
    \vspace{0.2cm} % Adds a little vertical space
    \small
    $^1$Shanghai AI Laboratory \quad
    \small
    $^2$The Chinese University of Hong Kong \\ [-0.5em]
    \small
    $^3$State Key Laboratory of Multimedia Information Processing, School of Computer Science, Peking University\\
    \small
    $^4$National Engineering Research Center of Visual Technology, School of Computer Science, Peking University \\
    \small 
    \small % 先把字号变小
    \texttt{photon@stu.pku.edu.cn, zhangfan@pjlab.org.cn, shiboxin@pku.edu.cn} \\
    \small
    \texttt{tfxue@ie.cuhk.edu.hk, wangyujin@pjlab.org.cn}
}

\begin{document}
\twocolumn[{%
\renewcommand\twocolumn[1][]{#1}%
\maketitle

\vspace{-15pt}
\begin{center}
\centering
\includegraphics[width=\textwidth]{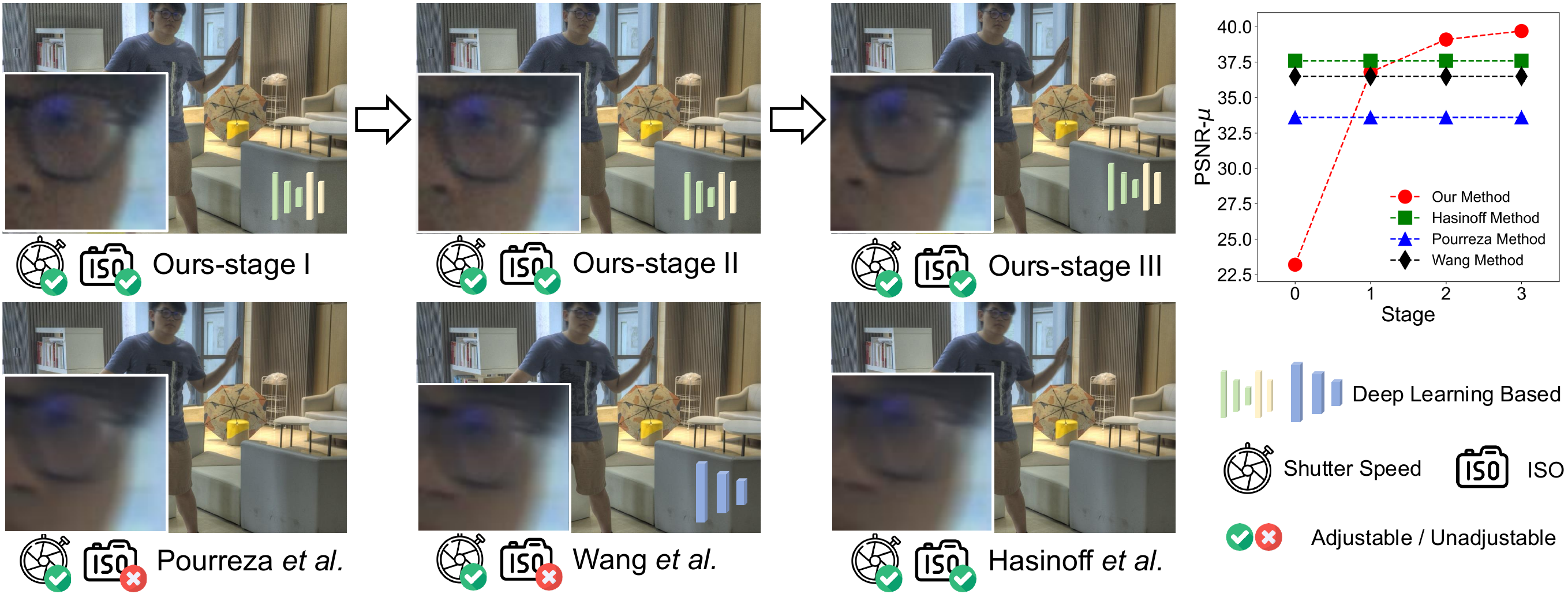}
\captionof{figure}{AdaptiveAE takes camera preview images as input and automatically predicts the ISO and shutter speed for each LDR captures for exposure fusion through a 3-stage sequential refinement procedure to achieve an optimal balance between noise level and motion-related problems for high quality HDR capturing in dynamic scenes with deep reinforcement learning. AdaptiveAE achieves PSNR 39.7 on HDRV dataset~\cite{hdrv}, while baseline methods~\cite{wang2020learning, hasinoff2010noise, pourreza2015exposure} that either only predicts shutter speed or do not consider motion can only achieve PSNR below 37.6 and has evident motion blur and ghosting artifacts in HDR results.}
\label{fig:teaser}
\end{center}

}]
\begingroup 
\renewcommand{\thefootnote}{\fnsymbol{footnote}} 
\footnotetext[1]{This work was done during Tianyi Xu's internship at Shanghai AI Laboratory.} 
\footnotetext[2]{Corresponding authors.}
\endgroup

\begin{abstract}

Mainstream high dynamic range imaging techniques typically rely on fusing multiple images captured with different exposure setups (shutter speed and ISO). A good balance between shutter speed and ISO is crucial for achieving high-quality HDR, as high ISO values introduce significant noise, while long shutter speeds can lead to noticeable motion blur. However, existing methods often overlook the complex interaction between shutter speed and ISO and fail to account for motion blur effects in dynamic scenes.

In this work, we propose AdaptiveAE, a reinforcement learning-based method that optimizes the selection of shutter speed and ISO combinations to maximize HDR reconstruction quality in dynamic environments. AdaptiveAE integrates an image synthesis pipeline that incorporates motion blur and noise simulation into our training procedure, leveraging semantic information and exposure histograms. It can adaptively select optimal ISO and shutter speed sequences based on a user-defined exposure time budget, and find a better exposure schedule than traditional solutions. Experimental results across multiple datasets demonstrate that it achieves the state-of-the-art performance.
\vspace{-15pt}
\end{abstract}

\section{Introduction}
\label{sec:intro}

High-dynamic-range (HDR) imaging plays a pivotal role in computational photography. To capture an HDR scene, due to hardware limits, a single capture can only cover a Low Dynamic Range (LDR), and HDR fusion techniques are proposed to combine multiple LDR images with varying exposures to cover a wide dynamic range~\cite{mertens2009exposure, ma2019deep, huang2022real,niu2021hdr-gan, liu2022ghost-hdr-transformer, qu2022transmef, kong2025safnet, zhang2023self, yan2020deep, yan2019attention}. The typical way to vary exposure is to change either shutter speed or ISO. This is a challenging process, as a longer shutter speed may increase the signal-to-noise ratio but introduce unpleasant motion blur, a large ISO may increase brightness but also magnify noise, and significant exposure differences can cover a wider dynamic range but also increase the risk of misalignment. Therefore, the choice of exposure values (EVs) for each capture is critical in this process to ensure high-quality HDR results.

Still, very limited work discusses how to choose the optimal exposure levels, particularly in dynamic scenes. Prior works on exposure scheduling mainly focus on static scenes and ignores potential motion blur. Learning-based techniques overlook the intricate interplay between ISO and shutter speed, resulting in suboptimal image quality under varying conditions~\cite{wang2020learning, pourreza2015exposure, hasinoff2010noise}. Additionally, many existing methods treat ghosting and motion blur as separate, computationally intensive post-processing tasks, which is unsuitable for real-time applications~\cite{kupyn2019deblurgan, tao2018scale, zhang2020deblurring, tsai2022banet, shen2019human, yuan2007image, vijay2013non}. 

In this work, we propose \emph{AdaptiveAE}, an efficient exposure control algorithm designed for HDR capturing, which addresses both motion blur and noise during image acquisition. Given the previously captured image, our method optimizes the exposure bracketing strategy for the subsequent capture, based on illumination information and semantic data from previous frames. Unlike previous approaches that treat motion blur as a separate post-processing task, we aim to address it during the capture process.

Designing both efficient and adaptive exposure control is non-trivial, and we resort to reinforcement learning~\cite{mnih2016asynchronous} to solve this challenge. Our solution mimics an experienced photographer. At each iteration, the policy network takes the previously captured LDR images as input, together with extracted semantic and illumination information. Given this information, the policy network learns to determine the optimal exposure setup for the following captures, which maximizes the additional information provided by this frame while also reducing the risk of misalignment and motion blur in a dynamic scene. Once the next frame is captured, this newly captured image will be used as input for the subsequent refinement iteration. As shown in \cref{fig:teaser} right, the final quality (PSNR) of fusion increases as more images are captured.

Our proposed approach offers several advantages over traditional exposure controls. First, \emph{AdaptiveAE} controls both exposure time and ISO and also adapts to different scenes. As a result, it has a much higher upper bound compared to either a fixed exposure schedule or an adaptive control algorithm that only changes exposure time. As shown in~\cref{fig:teaser}, our method iteratively achieves an optimal balance between noise and blur, resulting in less noise, reduced motion blur, and superior image quality compared to other baseline methods. Second, our method can handle both static and dynamic scenes. In static scenes, it achieves performance comparable to state-of-the-art techniques, and in dynamic ones, it produces visually compelling HDR images with minimal motion blur and ghosting artifacts. Third, our method can automatically choose the best number of frames for HDR imaging. Unlike traditional HDR approaches that often use a fixed number of frames, such as three, our approach provides flexibility to determine whether capturing three or more frames is optimal for certain scenes, balancing image quality and time budget.

Current datasets~\cite{chen2021hdr,hdrv,kalantari2017deep, kalantari2013patch, liu2023joint} are inadequate for studying auto-exposure (AE) with simultaneous noise and motion blur considerations. To bridge this gap, we introduce a blur-aware data synthesis pipeline. This novel approach enables the concurrent analysis of blur and noise in AE prediction, thereby enhancing HDR image quality. Our method uniquely integrates these factors, departing from traditional practices that address them separately.

We evaluate AdaptiveAE on established benchmarks, including the DeepHDR Video dataset~\cite{chen2021hdr} and the HDRV dataset~\cite{hdrv}, employing various downstream exposure fusion techniques. Our results demonstrate state-of-the-art performance compared to existing auto-exposure methods. Additionally, comprehensive ablation studies and targeted experiments focusing on motion blur confirm the efficacy of our approach and underscore the critical importance of incorporating blur synthesis into our pipeline. Our method, tested on real-world scenes using a SONY Alpha 7C-II, demonstrates superior noise control and effectively reduces motion blur, outperforming baselines in the visual quality of the fused HDR images.

% In summary, our contributions are threefold:
% \begin{itemize}
%     \item We introduced a blur-aware data synthesis pipeline for concurrent blur and noise analysis in AE prediction and, 
%     \item a new AE method that addresses motion blur and ghosting before LDR capturing, enhancing HDR quality.
%     \item We use deep reinforcement learning with adjustable ISO/exposures and custom rewards to optimize dynamic HDR exposure.
% \end{itemize}

Current datasets~\cite{chen2021hdr,hdrv,kalantari2017deep, kalantari2013patch, liu2023joint} are inadequate for studying auto-exposure (AE) with simultaneous noise and motion blur considerations. To bridge this gap, we introduce a blur-aware data synthesis pipeline. This novel approach allows for concurrent analysis of blur and noise in AE prediction, enhancing HDR image quality. Our method uniquely integrates these factors, departing from traditional practices that address them separately.

We evaluate AdaptiveAE on established benchmarks, including the DeepHDR Video dataset~\cite{chen2021hdr} and the HDRV dataset~\cite{hdrv}, employing various downstream exposure fusion techniques. Our results demonstrate state-of-the-art performance compared to existing auto-exposure methods. Also, comprehensive ablation studies and targeted experiments focusing on motion blur confirm the efficacy of our approach and highlight the critical importance of incorporating blur synthesis within our pipeline. Our method, tested on real-world scenes using a SONY Alpha 7C-II, demonstrates superior noise control and effectively reduces motion blur, outperforming compared baselines in the visual quality of the fused HDR images.
\vspace{-4pt}
\section{Related work}
\label{sec:related_work}

\noindent\textbf{Strategy for exposure bracketing.}
Determining the optimal set of exposures for multiple-exposure dynamic range imaging is a well-established problem. Most digital cameras allow users to set the compensation ratio for exposure bracketing, while mobile cameras typically impose fixed ratios during automatic exposure bracketing. Heuristic strategies based on the histogram are proposed by~\cite{gelfand2010multi,pourreza2015exposure,beekimproved}, to balance single-to-noise ratio (SNR) and saturation. The work by~\cite{hasinoff2010noise} firstly formulates this challenge as a constrained optimization problem in the linear RGB domain, addressing the scenario that follows multiple exposure fusion and precedes tone mapping and denoising. This concise formulation facilitates the use of a straightforward numerical solver. Further extensions of this formulation consider the alignment of multiple input images to account for handshake~\cite{seshadrinathan2012noise}. In the context of structured light 3D reconstruction, various exposures are likewise treated as an optimization problem~\cite{chen2022automated}. Exposure influences not only the noise but also, to some extent, the tone of the image in the standard RGB domain. Evaluating the final image quality complicates the problem further, as subsequent tone mapping or retouching can significantly alter image quality. Therefore, a neural network is utilized to estimate exposures and fuse multiple exposed images to achieve optimal fidelity in the gamma-corrected domain~\cite{li2023lightweight}. Additionally, reinforcement learning is employed by~\cite{yu2018deepexposure} and~\cite{wang2020learning} to assess the rewards on comprehensive image quality after more sophisticated tone adjustment.

\noindent\textbf{Challenge in dynamic scenes.}
In static scenes, increasing under-exposures can reduce saturation, and a longer shutter speed improves signal-to-noise ratio in dark areas. However, in dynamic scenes, excessive exposure may cause ghosting artifacts, and a prolonged shutter speed can lead to motion blur. Consequently, heuristic exposure bracketing is usually limited to two or three EV settings~\cite{gelfand2010multi,beekimproved}. The method by~\cite{hasinoff2010noise} imposes an upper limit on total shutter speed, while~\cite{seshadrinathan2012noise} addresses handshake motion through image registration but scarcely tackles local object motion. Most approaches do not fully address motion in dynamic scenes, leaving motion blur and ghosting artifacts to be addressed through post-processing.

\section{Method}
\label{sec:method}

\begin{figure}[tbp]
    \centering
    \begin{subfigure}{\linewidth}
        \centering
        \includegraphics[scale=0.39]{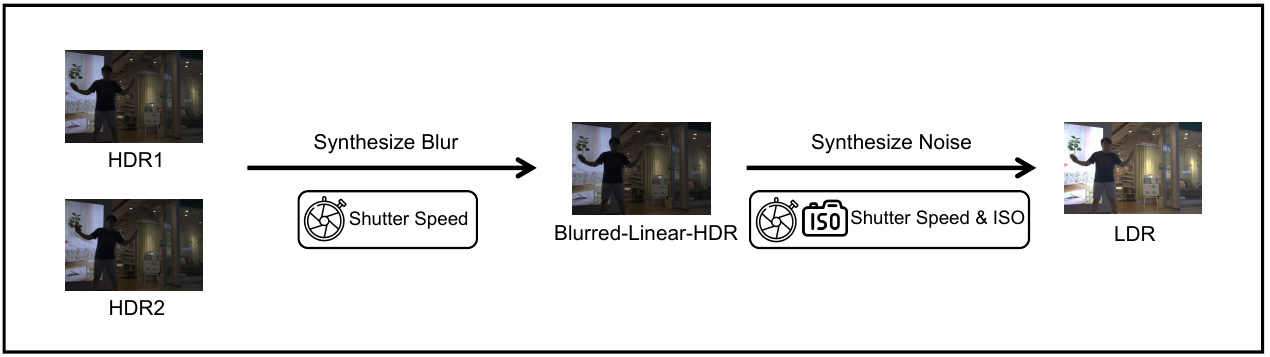}
        \caption{Our overall image synthesis pipeline.}
        \label{fig:synthesis_a}
    \end{subfigure}
    \hfill
    \begin{subfigure}{\linewidth}
        \includegraphics[scale=0.39]{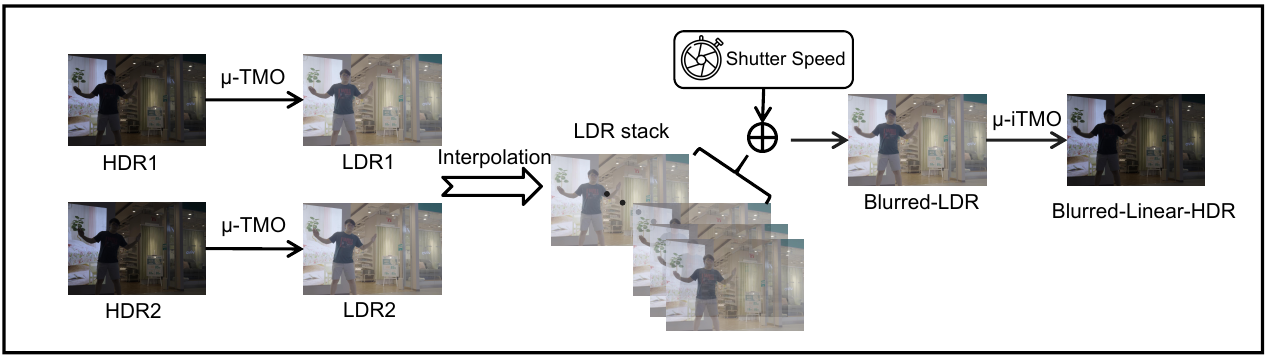}
        \caption{Our blur synthesis pipeline.}
        \label{fig:synthesis_b}
    \end{subfigure}
    \hfill
    \begin{subfigure}{\linewidth}
        \includegraphics[scale=0.39]{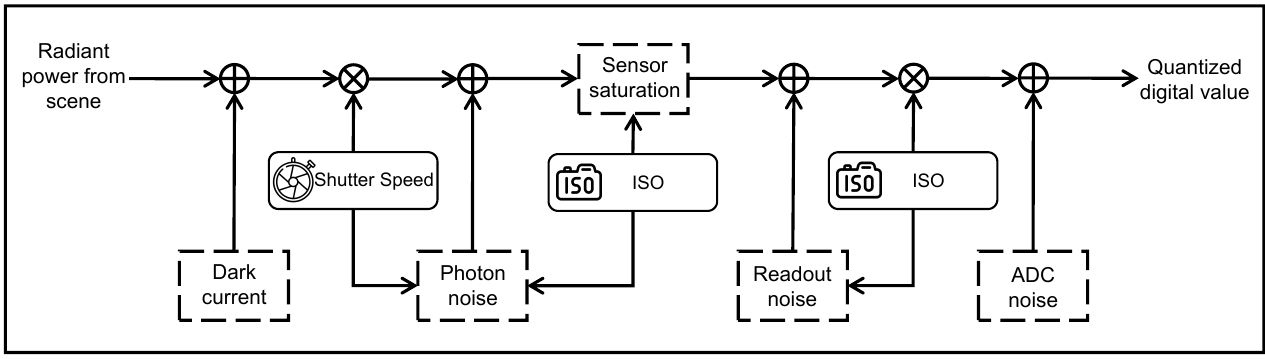}
        \caption{Our noise synthesis pipeline.}
        \label{fig:synthesis_c}
    \end{subfigure}
    \caption{Our blur-aware data synthesis pipeline.}
    \label{fig:synthesis}
\end{figure}

Conventional weighted linear combination methods, such as exposure fusion, provide straightforward SNR estimation but fail in dynamic scenes where moving objects create unquantifiable motion blur and ghosting artifacts due to misalignment. Our approach addresses these challenges during capturing rather than in post-processing, as the latter is shown to yield suboptimal results. We predict exposure-related risks—such as motion blur, ghosting, noise, and saturation—based on a limited number of previously captured frames. Additionally, we employ a sequential strategy for exposure and ISO parameter determination, rather than simultaneously predicting settings for all three LDR images, which reflects the iterative nature of auto-exposure in mobile cameras that enables adaptation to significant brightness transitions.

\subsection{Blur-aware data synthesis pipeline}
To simulate capturing in real environments, we designed an image synthesis pipeline to generate realistic motion blur and noise in LDR images from HDR videos in the training dataset, for use in training.
Typical exposure settings involve adjusting the exposure value (EV), which is calculated as:
\begin{equation}
    \text{EV} = \log_2 \left( \frac{F^2}{T} \times \frac{100}{\text{ISO}} \right), 
\label{ev}
\end{equation}
where $F$ denotes the aperture's f-number, \text{ISO} represents the ISO sensitivity and $T$ is the exposure time in seconds.

\begin{figure*}[tbp]
\centering
\includegraphics[scale=0.52]{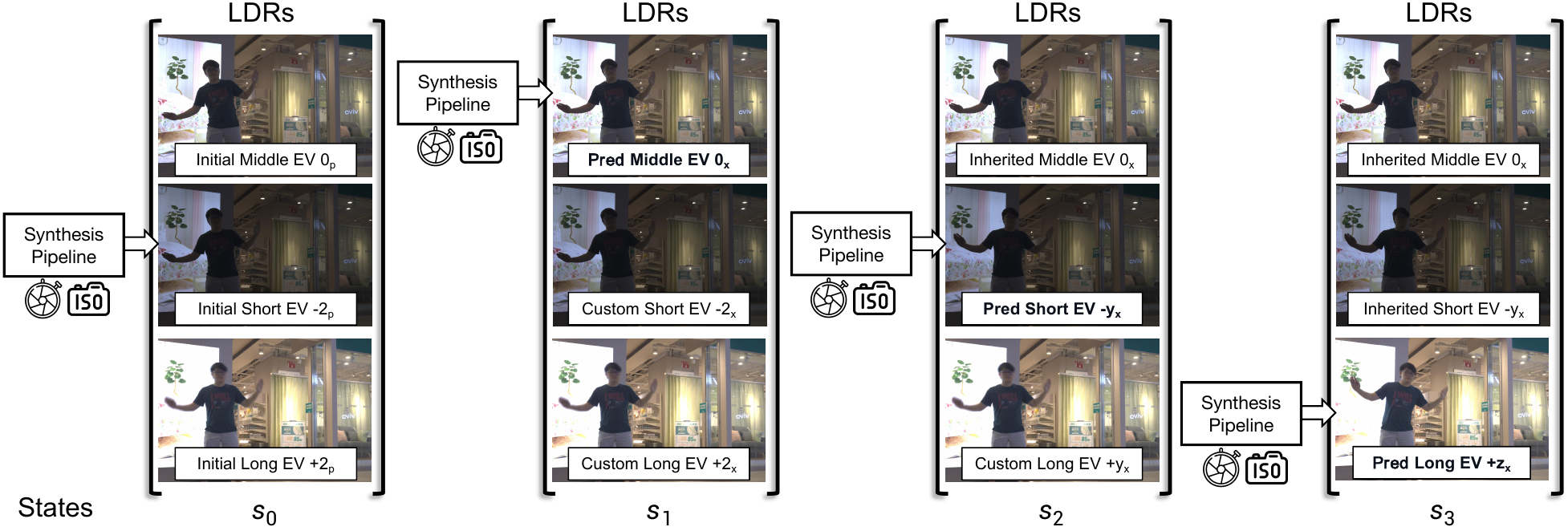}
\vspace{-5pt}
\caption{The training scheme of AdaptiveAE. States are defined as the three LDRs synthesized using predicted ISOs and shutter speeds. Starting from $s_0$ where the three LDRs has EV $\{-2,0,+2\}$ with arbitrary EV $0$ baseline, ISOs and shutter speeds, the agent sequentially predicts, customizes or inherits capturing parameters (\textit{i.e.} ISO and shutter speed) for the next stage and synthesize the corresponding LDR using our image synthesis pipeline. Unlike training, the LDRs will be captured rather than synthesized during inference.}
\label{side_pipeline}
\vspace{-15pt}
\end{figure*}
    
Similar to recent methods~\cite{wang2020learning, hasinoff2010noise, pourreza2015exposure} concerning high dynamic range capture, we assume that aperture and focus are held constant to prevent changes in defocus. This leaves just two camera settings to manipulate: (1) Shutter speed, which controls the amount of light to collect, and (2) ISO, which determines the sensor gain. 

% \vspace{-2pt}

    Our pipeline synthesizes motion blur and noise for the ground truth static HDR image based on a specified ISO and shutter speed. As depicted in \cref{fig:synthesis}, it uses two consecutive static HDR images as input. Motion blur is first synthesized according to shutter speed to produce a blurred HDR image in linear space. Next, noise is added based on shutter speed and ISO to create an LDR image, reflecting our exposure choice. Note that motion blur should be applied before adding noise, as it influences the raw input by affecting the number and pattern of captured photons during photography.

    \noindent\textbf{Synthesizing blur.} On-the-shelf training datasets consist of consecutive HDR ground truth of a scene with motion. Now we explain how we simulate motion blur to a frame of HDR $f_i^L$, as shown in~\cref{fig:synthesis_b}, which is the $i$-th HDR ground truth frame in the dataset scene, and the superscript $L$ indicates it is in linear space. We first use $\mu$-law tone-mapping with $\mu=5000$ to transfer $f_i^L$ and $f_{i+1}^L$ from HDR space to LDR space where the image interpolation algorithm we applied is trained upon, receiving $f_i^\mathcal{T}$ and $f_{i+1}^\mathcal{T}$, where the superscript $\mathcal{T}$ denotes they are in LDR space. Then we use RIFE~\cite{huang2022real} to interpolate the them to $256$ frames and get the sequence of images $\{ f_i^\mathcal{T}, s_1^\mathcal{T}, s_2^\mathcal{T}, \cdots ,s_{254}^\mathcal{T}, f_{i+1}^\mathcal{T} \}$. Then for the selected shutter speed $T_j$ for the $j$-th LDR $l_j^\mathcal{T}$ to take, the blurred HDR $b_j^L$ is simulated as:
\begin{equation}
    b_j^L = \mathbf{iTMO}(\frac{f_i^\mathcal{T}+\sum_{m=1}^{m_j} s_m^\mathcal{T}}{m_j}), m_j=\left\lceil \frac{256T_j}{\Delta \tau} \right\rceil,
\end{equation}
    where $\Delta \tau$ denotes the time elapsed from $f_i^L$ is taken to $f_{i+1}^L$ is not yet taken and $\mathbf{iTMO}$ indicates the inverse $\mu$-law tonemapping function with $\mu=5000$. 

    \noindent\textbf{Synthesizing noise.} We adopt the noise model mentioned in~\cite{hasinoff2010noise}, in which noise is modeled as a zero-mean variable, coming from three independent sources, including photon noise, which represents the Poisson distribution of photon arrivals and depends linearly upon the number of recorded electrons, $\Phi T$, readout noise, which comes from sensor readout, and analog-to-digital conversion(ADC) noise, which comes from the combined effect of amplifier and quantization. Hence, for pixels below the saturation level:
\begin{equation}
    Var(n) = \frac{\Phi T \times \text{ISO}^2}{U^2} + \frac{\sigma_{\text{read}}^2 \times \text{ISO}^2}{U^2} + \sigma_{\text{ADC}}^2,
\label{noise_equation_main}
\end{equation}
where $\Phi$ is the radiance level, $T$ is the shutter speed, $U$ is a camera-dependent variable.

    Following our noise model, as shown in~\cref{fig:synthesis_c}, we can synthesize the corresponding noise with selected ISO and shutter speed to the blurred HDR $b_j^L$ to get the LDR image $l_j^\mathcal{T}$. For details, please refer to our supplementary material.

\begin{figure*}[tbp]
\centering
\includegraphics[scale=0.52]{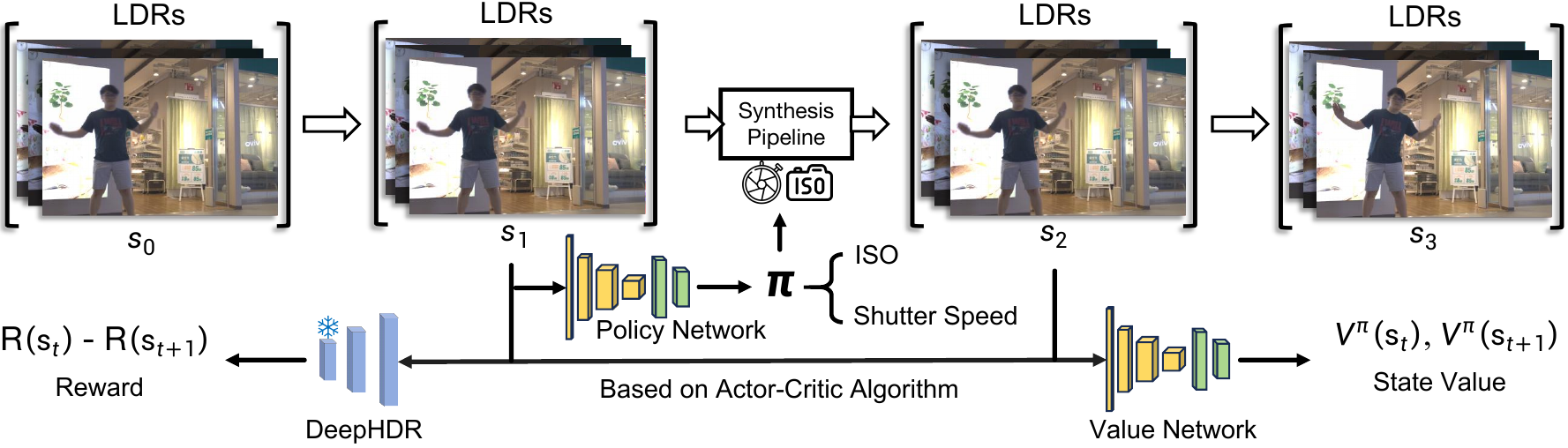}
\vspace{-5pt}
\caption{Training pipeline of our method. The ISO and shutter speed prediction process is conceptualized as a Markov Decision Process, where a CNN-based policy network predicts the ISO and shutter speed of the next exposure sets. Concurrently, a CNN-based value network estimates the state value. We leverage our blur-aware image synthesis pipeline to synthesize the predicted LDRs and employ DeepHDR~\cite{huang2022real} to fuse the predicted LDR images, generating our HDR result and calculating the reward for the current policy. The entire system is optimized using the A3C (Asynchronous Advantage Actor-Critic) method~\cite{mnih2016asynchronous}.}
\label{pipeline}
\vspace{-15pt}
\end{figure*}

% \subsection{Reinforcement learning}
\subsection{Problem formulation of AdaptiveAE}

    Given a scene, the goal of AdaptiveAE, which has access to three initial preview LDR images $\{p_j^\mathcal{T}\}_{1,2,3}$ (\textit{i.e.} underexposed, mid-exposed, overexposed) before capturing, is to find an optimal exposure setup (\textit{i.e.} ISO and shutter speed) for LDR capturing, with which the fused HDR output will result in pleasing visual performance. 

    Our method formulates exposure bracketing as a Markov Decision Process~\cite{puterman1990markov}, solved via deep reinforcement learning to refine exposure parameters (ISO, shutter speed) sequentially. As illustrated in \cref{pipeline}, the process starts from three LDRs at a default $\{-2_p, 0_p, +2_p\}$ EV spacing relative to an arbitrary reference (subscript $p$). The refinement then proceeds in stages, as shown in \cref{side_pipeline}.

    First, the agent predicts optimal parameters for the mid-exposed frame, establishing a new 0-EV reference (subscript $x$). The side frames are then \textbf{customized}—their parameters are procedurally set to achieve a symmetric $\{-2_x, 0_x, +2_x\}$ EV bracket. Next, the agent refines the underexposed frame to an EV of $-y_x$. The mid-exposed frame ($0_x$) is \textbf{inherited} (its parameters are reused), while the overexposed frame is customized to $+y_x$ to maintain symmetry, yielding $\{-y_x, 0_x, +y_x\}$. Finally, the agent predicts the overexposed frame's EV as $+z_x$, creating a potentially asymmetric set $\{-y_x, 0_x, +z_x\}$. This sequential prediction can be extended, as our adopted fusion method~\cite{huang2022real} handles more than three LDRs. A video providing further details on this process, along with an example of extended exposure bracketing, is available in the supplementary materials.

\subsection{Optimization objectives}

    Let us denote the problem as $P=(S, A)$, where S is a state space and A is an action space. Specifically, in our task, $S$ is the space of the exposure setups (\textit{i.e.}, ISO and shutter speed) for an LDR set that typically contains three LDRs (\textit{i.e.} under-exposed, mid-exposed, and over-exposed), while $A$ is the set of all possible ISO and shutter speed combinations, which is discrete. During training, at stage $s_j=\{ (\text{ISO}_{j1}, T_{j1}),(\text{ISO}_{j2}, T_{j2}),(\text{ISO}_{j3}, T_{j3}),  \}$, we first find the corresponding HDR ground truth image pairs $(f_i^L, f_{i+1}^L)_{1,2,3}$ and generate the corresponding LDRs $\{l_j^\mathcal{T}\}_{1,2,3}$ through our image synthesis pipeline in ~\cref{fig:synthesis}. The footnote ${1,2,3}$ denotes that the same operation is done for the under-, mid-, and over-exposed LDRs. Taking $\{l_j^\mathcal{T}\}_{1,2,3}$ as input, the agent predicts an action $a_j=(\text{ISO}_j, T_j)$, which is expanded to an exposure setup for three LDRs by customizing EV or inheriting from state $s_{j-1}$, mapping state $s_j$ to state $s_{j+1}$. Adding a sequence of $M$ LDRs to exposure bracketing corresponds to a trajectory $\tau$ of states and actions:
\begin{equation}
    \tau = (s_0, a_0, \cdots , s_{M-1}, a_{M-1}, s_M),
\end{equation}
    where $s_M$ is the stopping state. Our goal is to find a policy that maximizes the accumulated reward during the decision-making process. In this paper, the reward function with the $j$-th action (i.e., corresponding to deciding the exposure setting for the $j$-th LDR for exposure fusion) is thus written as:
\begin{equation}
    r(s_j, a_j)=\mathcal{R}(s_{j+1})-\mathcal{R}(s_j)-\mathcal{P}(j),
\end{equation}
    where $s_{j+1}=p(s_j, a_j)$, and $\mathcal{R}$ denotes our reward design and $\mathcal{P}$ denotes the $L_{\text{step}}$ penalty, detailed in \ref{reward_setting}.

    As depicted in \cref{pipeline}, our model comprises a policy network and a value network, both of which utilize a CNN-based architecture. 
    The policy network predicts the optimal ISO and shutter speed for the subsequent exposure, outputting a distribution of action probabilities $\pi(s, \theta)$ for an input image $s$. 
    Concurrently, the value network $V^{\pi}(s, \omega)$ estimates the corresponding state value. 
    These networks, with combined parameters $\psi=(\theta, \omega)$, are trained by maximizing our objective $J(\theta)_{\psi}$ to learn the optimal policy $\pi(s)$. Specifically, to train the policy network and the value network, we apply the A3C (Asynchronous Advantage Actor-Critic) method~\cite{mnih2016asynchronous}, where the actor is represented by the policy network and the critic is the value network. Network details are in the supplementary materials.
    % When selecting the $j$-th pair of camera settings $(\text{ISO}_j,T_j)$, the input of our policy network is a low-resolution LDR preview image $s_j^\mathcal{T}$ and the fused HDR $h_j^L$ of the LDRs in the LDR stack $\{ l_1^\mathcal{T}, l_2^\mathcal{T}, \cdots ,l_{j-1}^\mathcal{T} \}$, which contains the LDRs that are already taken with our predicted camera settings. To sum up, the input to the policy network is $(s_j^\mathcal{T}, h_j^L)$.
    
    % In order to further perceive the input image and get a better latent embedding for learning, we utilize two branches for feature extraction. 
    
    % \noindent\textbf{Semantic Branch.} Following the design of ~\cite{wang2020learning}, we use Alexnet~\cite{krizhevsky2012imagenet} to extract the semantic feature of the LDR preview image $s_j^\mathcal{T}$.

    % \noindent\textbf{Illumination Branch.} Still following the design of ~\cite{wang2020learning}, we estimate the histogram of the current HDR result $h_j^L$ with a 3-level spatial pyramid.
    
    % We further use a densely connected layer to fuse the features extracted by both branches to receive the final scene feature, which will be the input to our policy network. Detailed network design is explained in our supplementary material.

\subsection{Reward}
\label{reward_setting}

    When designing the reward function for our system, we considered four key factors: (1) similarity between the fused HDR and the ground truth HDR; (2) quality of important regions in the fused HDR; (3) quality of moving regions in the fused HDR; and (4) a penalty for overly long LDR stacks.
    Thus, our reward function is:
    \begin{equation}
        \mathcal{R} = -(P_{\text{construction}}+P_{\text{priority}}+P_{\text{ghost}}).
    \end{equation}
    We consider $P_{\text{construction}}$, the L2 loss between our fused HDR and the ground truth, as our major reward component, which is affected by noise and saturation. Note that through the entire sequential decision process, the middle-exposed frame is used as a reference for HDR fusion. Conforming to \cref{noise_equation_main}, during training, noise is synthesized according to the irradiance of ground truth HDR, and during inference, noise is estimated with the irradiance of noisy signals.
    
    $P_{\text{priority}}$ represents an L2 loss within the areas in the image masked by an importance mask, which is generated by a saliency predictor~\cite{pan2017salgan}. This ensures that the highest quality is maintained in the most significant areas, thereby enhancing the overall visual fidelity where it matters most. 

    $P_{\text{ghost}}$ is also an L2 loss within a masked area, denoting areas with large motions and thus having a higher risk of motion blur- or ghosting-caused HDR quality degradation. The mask is computed by calculating the optical flow using RAFT~\cite{teed2020raft} between the HDR ground truth of the middle-exposed frame (i.e., the reference frame for fusion) and the corresponding HDR $f_i^L$ and selecting the pixels where the mode of the flow vector exceeds a constant threshold $K$. Normalizing the largest optical flow vector, we empirically set $ K$ to 0.2. $P_\text{ghost}$ guides the agent to deal more carefully with regions that are prone to artifacts caused by motion and is helpful for high-quality HDR capturing, as is verified by the result of our ablation studies in \cref{ablation}.

    $\mathcal{P}(j)$ is a penalty designed to penalize excessively long exposure brackets. Capturing an excessive number of shots, such as 10, even with random exposure settings, can lead to nearly perfect HDR fusion results, but it is time-consuming. Typically, three shots~\cite{barakat2008minimal, pourreza2015exposure, hasinoff2010noise} are sufficient to achieve high-quality outcomes. To this end, we incorporate a penalty for taking more than three shots, as follows:
\begin{equation}
    \mathcal{P}(j) = 
    \begin{cases}
    0 & \text{if } j \le H \\
    \alpha (j-H)^2 & \text{if } j > H
    \end{cases},
\label{step_penalty}
\end{equation}
    where $\alpha$ is a positive coefficient and $H$ is set to 3. 

    In this manner, the autonomous agent optimizes exposure parameters by predicting relatively fast shutter speeds for LDR images, particularly for the middle-exposed reference frame, thereby minimizing motion blur while avoiding excessive ISO values that would introduce noise-related degradation. When confronted with potential ghosting artifacts—which emerge from information deficiency in LDR images due to concurrent saturation and motion—the agent adaptively selects EVs that minimize both underexposure and saturation, resulting in a significant reduction of ghosting artifacts in the final reconstruction.

\section{Experiments}
\label{sec:experiments}

\begin{figure*}[tbp]
\centering
\includegraphics[scale=0.3]{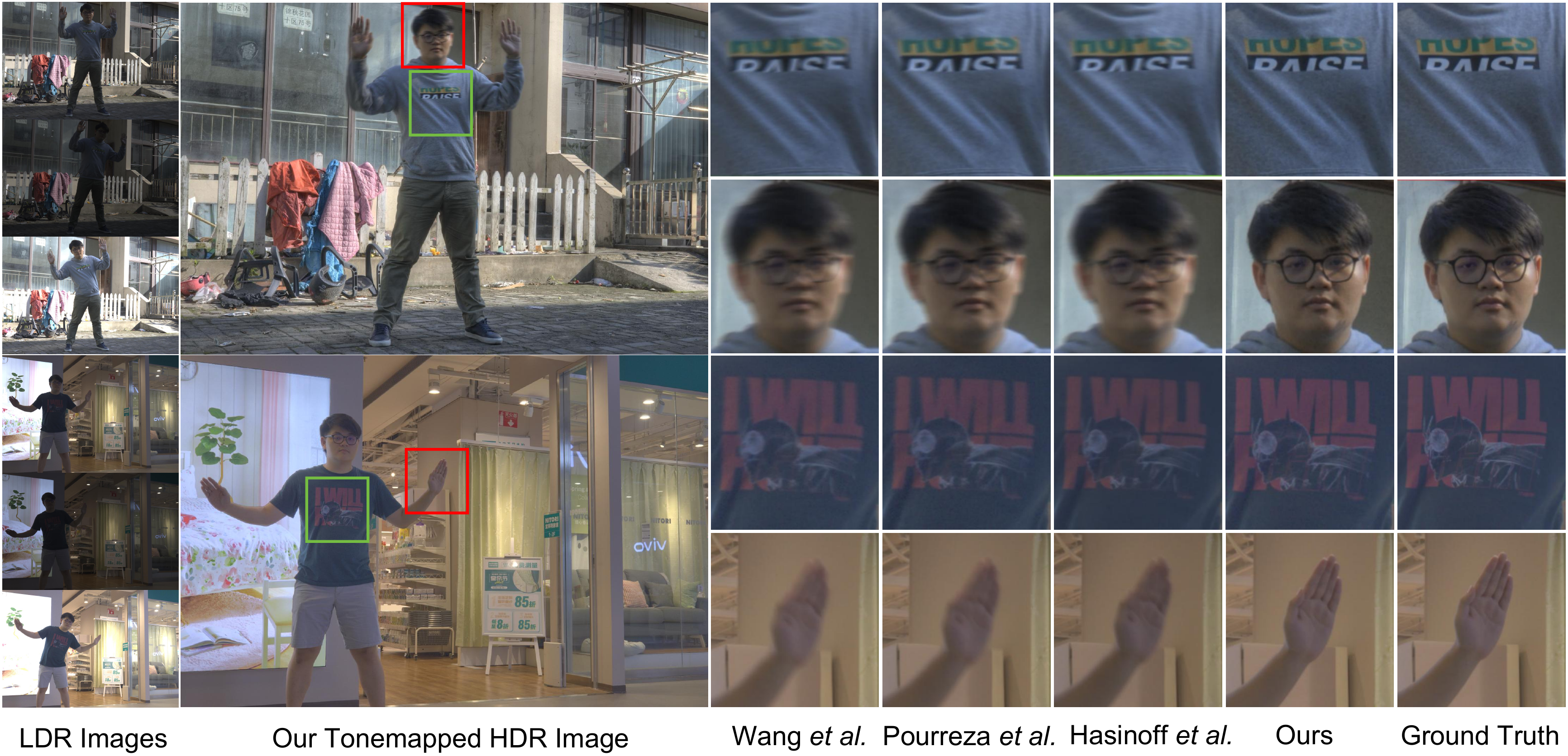}
\vspace{-5pt}
\caption{Qualitative comparisons with other auto-exposure methods on HDRV dataset~\cite{hdrv}. Left: Predicted LDRs with varying ISO and shutter speed settings and synthesized using our image synthesis pipeline. Middle: Fused HDR image using DeepHDR~\cite{huang2022real} and tone-mapped using Photomatix. Right: Zoom-in results for tested methods.}
\label{main_display}
\vspace{-4pt}
\end{figure*}

% \subsection{Experiment Settings}
% \noindent\textbf{Dataset.}
%     We use Real-HDRV~\cite{hdrv} for training and tested our performance on Real-HDRV~\cite{hdrv} and DeepHDRVideo~\cite{chen2021hdr}. For methods that do not account for changes in ISO values, we set the ISO to 200 as a standard value for most cases. This practice is rationalized in our supplementary materials.
    
\noindent\textbf{Experiment details.}
    We use Real-HDRV~\cite{hdrv} for training and tested our performance on Real-HDRV~\cite{hdrv} and DeepHDRVideo~\cite{chen2021hdr}. For HDR fusion, we adopt DeepHDR~\cite{wu2018deep} to generate the HDR image based on the selected exposure bracketing. Additionally, we set the number of LDR frames to 3 for all methods involved. For methods that do not account for changes in ISO values, we set the ISO to 200 as a standard value for most cases, which is rationalized in our supplementary materials.
    Since RIFE~\cite{huang2022real} is relatively time-consuming, the image interpolation and blur synthesis step is performed before training.
    We apply random flipping, rotation, and cropping with 512 $\times$ 512 pixels for data augmentation. Our training dataset consists of a total of 770 scenes, including 440 dynamic and 330 static scenes.
    
\noindent\textbf{Evaluation metrics. }
    We directly evaluate the performance of the fused HDR results. Similar to previous HDR fusion methods~\cite{liu2022ghost-hdr-transformer, niu2021hdr-gan, huang2022real}, we employ PSNR-$\mu$, SSIM-$\mu$, PU-PSNR, PU-SSIM, and HDR-VDP-2~\cite{mantiuk2011hdr_vdp_2} as evaluation metrics. PSNR-$\mu$ and SSIM-$\mu$ denote PSNR and SSIM of the fused HDR after $\mu$-law tone-mapping with $\mu$=5000. PU-PSNR and PU-SSIM are computed after perceptually uniform encoding~\cite{azimi2021pu21}. When computing the HDR-VDP-2 ~\cite{mantiuk2011hdr_vdp_2}, the diagonal display size is 30 inches.
    % Since the goal of our trained agent is not achieving SNR-optimal capture like Wang~\etal~\cite{wang2020learning} or Hasinoff~\etal~\cite{hasinoff2010noise} but striking a balance between noise and artifacts caused by movement in the scene to achieve the highest visual HDR quality, we leverage a variety of metrics to validate the effectiveness of our model, including PSNR-$\mu$, SSIM-$\mu$, HDR-VDP-2~\cite{mantiuk2011hdr_vdp_2}, PU-PSNR and PU-SSIM. PSNR-$\mu$ and SSIM-$\mu$ denote PSNR and SSIM of the fused HDR after $\mu$-law tone-mapping with $\mu$=5000. PU-PSNR and PU-SSIM are computed after perceptually uniform encoding~\cite{azimi2021pu21}. When computing the HDR-VDP-2 ~\cite{mantiuk2011hdr_vdp_2}, the diagonal display size is 30 inches.

\subsection{Results}
% \noindent\textbf{Results on Real-HDRV Dataset~\cite{hdrv} and DeepHDRVideo Dataset~\cite{chen2021hdr}.}
\noindent\textbf{Results on Real-HDRV dataset.}
    We compared our trained agent's performance with several state-of-the-art HDR exposure bracketing methods, including Pourreza-Shahri~\etal~\cite{pourreza2015exposure}, Hasinoff~\etal~\cite{hasinoff2010noise}, and Wang~\etal~\cite{wang2020learning}. The first two are non-deep-learning methods that do not consider motion: Pourreza-Shahri~\etal. use K-means clustering to adjust shutter speed based on image brightness, while Hasinoff~\etal. mathematically optimize ISO and shutter speed for the best worst-case SNR. Wang~\etal utilize reinforcement learning to predict shutter speed for maximizing PSNR, but also ignore motion.

    Our method achieves state-of-the-art performance on the HDRV-Test dataset, as shown in \cref{table_methods}. By treating ISO as a variable, it provides flexibility in handling extremely dark scenes. Additionally, it excels in dynamic scenes due to a blur synthesis model and carefully designed rewards. Thus, our model effectively balances noise reduction and motion artifact minimization, delivering high-quality HDR results. Visualization results in \cref{main_display} show that deep learning-based exposure fusion models affect EV selection differently. In Hasinoff~\etal's setup, three LDRs are equally weighted, but dynamic scenes require a reference image, emphasizing the middle-exposed LDR's quality. This often shifts its EV toward under-exposure to reduce motion blur. Our experiments confirm that if the middle-exposed image is blurry—a common problem in other methods—the fused HDR will also be blurred.

\begin{table*}[tbp]
\centering
\small
\caption{Comparison of Different Methods. We utilized one preview image and three preview images, respectively, for the compared methods and our method to predict exposure settings for three LDRs. We leveraged DeepHDR~\cite{huang2022real} for exposure fusion and used the mentioned metrics to evaluate the quality of the fused HDR. \textbf{Bold}: The best.}
\vspace{-5pt}
\resizebox{\textwidth}{!}{% Resizes table to fit textwidth
\begin{tabular}{c|ccccc|ccccc}
\toprule
\multirow{2}{*}{Methods} & \multicolumn{5}{c|}{HDRV \cite{hdrv}} & \multicolumn{5}{c}{DeepHDRVideo \cite{chen2021hdr}} \\
\cmidrule(lr){2-6} \cmidrule(lr){7-11}
& PSNR-$\mu$ & SSIM-$\mu$ & HDR-VDP-2 & PU-PSNR & PU-SSIM & PSNR-$\mu$ & SSIM-$\mu$ & HDR-VDP-2 & PU-PSNR & PU-SSIM \\
\midrule
Pourreza \etal \cite{pourreza2015exposure} & 33.64 & 0.8617 & 54.55 & 30.61 & 0.8679 & 35.57 & 0.8780 & 55.67 & 31.59 & 0.8791 \\
Hasinoff \etal \cite{hasinoff2010noise} & 37.59 & 0.9052 & 57.02 & 32.87 & 0.8980 & 38.47 & 0.9157 & 58.65 & 34.45 & 0.9132 \\
Wang \etal \cite{wang2020learning} & 36.46 & 0.8902 & 56.09 & 32.68 & 0.8933 & 37.95 & 0.9019 & 57.39 & 33.27 & 0.9008 \\
Ours & \textbf{39.70} & \textbf{0.9408} & \textbf{59.20} & \textbf{34.67} & \textbf{0.9465} & \textbf{39.81} & \textbf{0.9371} & \textbf{58.90} & \textbf{36.19} & \textbf{0.9338} \\
\bottomrule
\end{tabular}
} % End of resizebox
\label{table_methods}
\vspace{-15pt}
\end{table*}

% \begin{table*}[htbp]
% \centering
% \caption{Comparison of Different Methods}
% \resizebox{\textwidth}{!}{%  <--- Resizes table to fit textwidth
% \begin{tabular}{lcccccccccc}
% \toprule
% \multirow{2}{*}{Methods} & \multicolumn{5}{c}{Dynamic set} & \multicolumn{5}{c}{Static set} \\
% \cmidrule(lr){2-6} \cmidrule(lr){7-11}
% & PSNR-$\mu$ & SSIM-$\mu$ & HDR-VDP-2 & PU-PSNR & PU-SSIM & PSNR-$\mu$ & SSIM-$\mu$ & HDR-VDP-2 & PU-PSNR & PU-SSIM\\
% \midrule
% Pourreza & 30.39 & 0.8257 & 52.20 & 27.66 & 0.8320 & 37.98 & 0.9142 & 57.68 & 34.54 & 0.9158 \\
% Hasinoff & 33.15 & 0.8572 & 54.17 & 29.16 & 0.8478 & \textbf{43.52} & \textbf{0.9691} & \textbf{60.83} & \textbf{37.82} & \textbf{0.9650} \\
% Wang    & 32.03 & 0.8431 & 53.26 & 28.88 & 0.8459 & 42.36 & 0.9529 & 59.87 & 37.75 & 0.9566 \\
% Ours w/o blur & 32.76 & 0.8466 & 53.28 & 28.76 & 0.8500 & 42.96 & 0.9603 & 60.32 & 37.28 & 0.9614 \\
% Ours    & \textbf{37.62} & \textbf{0.9249} & \textbf{58.01} & \textbf{32.41} & \textbf{0.9358} & 42.47 & 0.9621 & 60.78 & 37.69 & 0.9607 \\
% \bottomrule
% \end{tabular}
% } % <--- End of resizebox
% \label{table_methods}
% \end{table*}

\begin{table}[tbp]
\centering
\small
\caption{Ablation study of AdaptiveAE on HDRV~\cite{hdrv} dataset. Base denotes our model trained with only the step penalty and construction reward. \textbf{Bold}: The best.}
\setlength{\tabcolsep}{1pt} % Adjust this value as needed
\begin{tabular}{l|ccccc}
\hline
Model & PSNR-$\mu$ & SSIM-$\mu$ & PU-PSNR & PU-SSIM & \\ \hline
Base & 38.21 & 0.9227 & 32.68 & 0.9198 \\
Base+$P_{\text{priority}}$ & 38.57 & 0.9261 & 33.02 & 0.9239 \\
Base+$P_{\text{priority}}$+$P_{\text{ghost}}$ & \textbf{39.70} & \textbf{0.9408} & \textbf{34.67} & \textbf{0.9465} \\ \hline
\end{tabular}
\label{ablation}
\vspace{-15pt}
\end{table}

% \begin{table*}[h!]
% \centering
% \caption{Generalization tests on Chen\cite{chen2021hdr}}
% \resizebox{\textwidth}{!}{%  <--- Resizes table to fit textwidth
% \begin{tabular}{lcccccccccc}
% \toprule
% \multirow{2}{*}{Methods} & \multicolumn{5}{c}{Dynamic set} & \multicolumn{5}{c}{Static set} \\
% \cmidrule(lr){2-6} \cmidrule(lr){7-11}
% & PSNR-$\mu$ & SSIM-$\mu$ & HDR-VDP-2 & PU-PSNR & PU-SSIM & PSNR-$\mu$ & SSIM-$\mu$ & HDR-VDP-2 & PU-PSNR & PU-SSIM\\
% \midrule
% Pourreza & 32.10 & 0.8453 & 53.65 & 29.64 & 0.8485 & 40.20 & 0.9216 & 58.37 & 34.19 & 0.9198 \\
% Hasinoff & 34.78 & 0.8826 & 56.21 & 31.39 & 0.8803 & \textbf{43.38} & \textbf{0.9599} & \textbf{61.90} & \textbf{38.52} & \textbf{0.9570} \\
% Wang    & 34.35 & 0.8695 & 55.36 & 29.96 & 0.8653 & 42.76 & 0.9450 & 60.10 & 37.69 & 0.9482 \\
% Ours w/o blur & 33.82 & 0.8777 & 55.23 & 30.71 & 0.8694 & 43.09 & 0.9597 & 60.48 & 38.15 & 0.9536 \\
% Ours    & \textbf{37.21} & \textbf{0.9250} & \textbf{58.17} & \textbf{34.80} & \textbf{0.9187} & 43.28 & 0.9533 & 59.87 & 38.04 & 0.9540 \\
% \bottomrule
% \end{tabular}
% } % <--- End of resizebox
% \label{table_chen}
% \end{table*}

\begin{table}
\centering
\small
\caption{Comparison of different exposure fusion methods. We utilize the four auto-exposure methods to predict exposure settings and employ our image synthesis pipeline to create three LDRs. Then, we apply different exposure fusion methods to fuse them and compare the HDR quality on the HDRV dataset~\cite {hdrv}. HDR-Transformer~\cite{liu2022ghost-hdr-transformer} is pretrained on HDRV~\cite{hdrv} dataset. P: PSNR-$\mu$, S: SSIM-$\mu$. Pou: Pourreza \etal \cite{pourreza2015exposure}, Has: Hasinoff \etal \cite{hasinoff2010noise}, W: Wang \etal \cite{wang2020learning}. HDR-Trans: HDR-Transformer. \textbf{Bold}: The best.}
\label{tab:hdr_comparison}
\begin{tabular}{l|cc|cc|cc}
\hline
\multirow{2}{*}{Model} & \multicolumn{2}{c|}{DeepHDR} & \multicolumn{2}{c|}{HDR-GAN} & \multicolumn{2}{c}{HDR-Trans} \\
& P-$\mu$ & S-$\mu$ & P-$\mu$ & S-$\mu$ & P-$\mu$ & S-$\mu$ \\
\hline
Pou & 33.64 & 0.8617 & 35.71 & 0.8892 & 35.84 & 0.8824 \\
Has & 37.59 & 0.9252 & 38.58 & 0.9263 & 39.11 & 0.9372 \\
W & 36.46 & 0.9002 & 37.95 & 0.9169 & 38.89 & 0.9210 \\
Ours & \textbf{39.70} & \textbf{0.9408} & \textbf{40.73} & \textbf{0.9376} & \textbf{41.37} & \textbf{0.9478} \\
\hline
\end{tabular}
\label{different_fusion}
\vspace{-10pt}
\end{table}

\noindent\textbf{Inference time.} 
    Our method takes less than 250 ms for each HDR image, which is acceptable. With average exposure time $n$ $(\le 30 ms)$ and prediction time $m$ $(\le 10 ms)$, the total execution time is $6n + 3m$ $(\le 250ms)$. This can be further lowered to $6n$ $(\le 200 ms)$ adopting concurrency. Note that DeepHDR is only required during training for calculating rewards, and on-camera inference can be performed with a small agent, taking only $3.5ms$ for each frame without optimization. Besides, our method can be further accelerated by on-camera processors when applied to DSLRs. More discussions are in our supplementary materials.

\begin{table}[tbp] 
\centering 
\caption{Performance comparison investigating the gap between our method and the optimum on the HDRV-test dataset.} 
\label{tab:gap}
\begin{tabular}{lcccc} 
\hline
           & Ours & Worst & Average & Best \\ 
\hline
PSNR-$\mu$ & 39.70    & 25.76     & 32.41       & 39.93    \\ 
SSIM-$\mu$ & 0.9408    & 0.7738     & 0.8609       & 0.9412    \\ 
\hline
\vspace{-5pt}
\end{tabular}
\end{table}

\noindent\textbf{Gap to the best-achievable.}
    For each scene in the test set of Real-HDRV, we iteratively search for the best set of predictions by Gaussian sampling around our initial prediction (50 times per exposure parameter per frame, with a deviation of 20\% of the mean for both ISO and shutter speed). Statistics in \cref{tab:gap} show that our method approaches the locally optimal result while being efficient.

\noindent\textbf{Cross datasets test.}
    To test the generalization ability of our model, we also evaluated the performance of our trained agent on DeepHDRVideo~\cite{chen2021hdr}, as shown in the right of ~\cref{table_methods}. Our agent exhibits good generalization abilities. Since the DeepHDRVideo dataset only provides ground truth HDR for the middle image in the sequence, we synthesize HDR for the other images using DeepHDR~\cite{wu2018deep}.

% \begin{figure}[tbp]
% \centering
% \includegraphics[scale=0.39]{img/generalization.pdf}
% \vspace{-5pt}
% \caption{Results on DeepHDRVideo~\cite{chen2021hdr} dataset. Ground truth is synthesized by using DeepHDR~\cite{huang2022real} to merge the provided LDRs. \vspace{-5pt}}
% \label{chen_result}
% \end{figure}

\noindent\textbf{Cross HDR fusion methods test.} 
    In our training scheme, we used DeepHDR for exposure fusion. For further testing, we adopted different exposure methods, including HDR-GAN~\cite{niu2021hdr-gan} and HDR-Transformer \cite{liu2022ghost-hdr-transformer} (pre-trained on the HDRV-dataset~\cite{hdrv}), as post-processing exposure fusion methods. All tests are conducted on the HDRV-Test dataset. As shown in \cref{tab:hdr_comparison}, without considering motion before LDR capturing, however powerful an exposure fusion method (\textit{i.e.} HDR-GAN and HDR-Transformer) fails to yield a satisfactory result. Notably, when stronger fusion models are used, the performance gap between our model and traditional models, which do not account for motion, increases. This is because, when blur and ghosting risk are mitigated, post-processing models can effectively manage the remaining challenges. In contrast, ignoring motion makes these challenges more difficult for post-processing models, leading to potential failure.

\begin{figure}[tbp]
\centering
\includegraphics[scale=0.39]{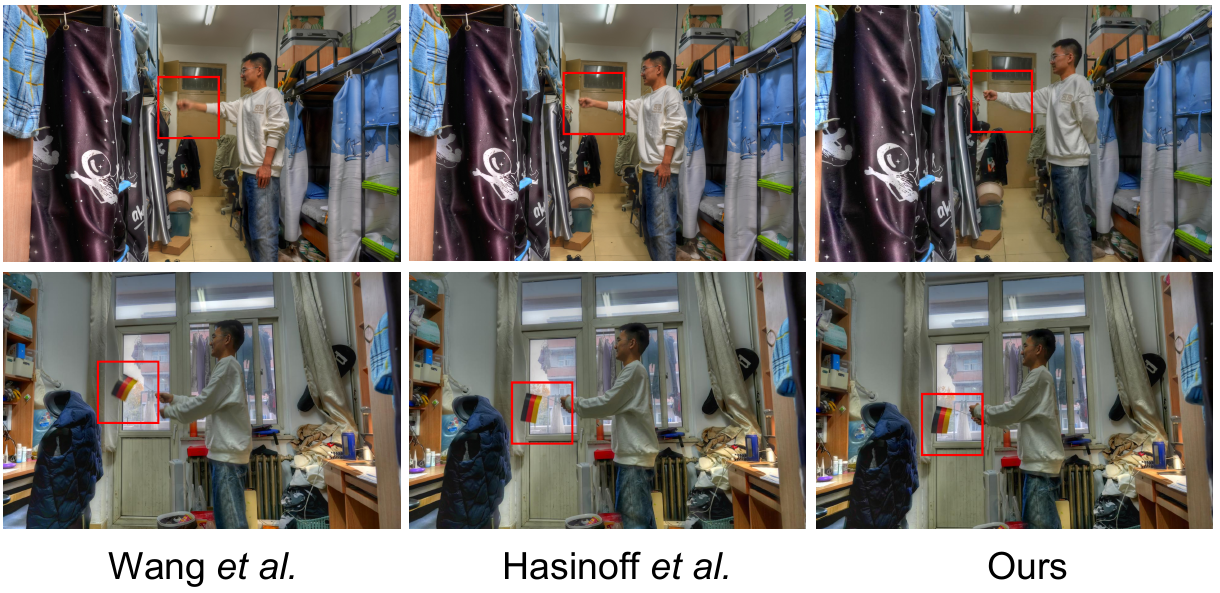}
\vspace{-3pt}
\caption{Results of our real capture data. The subject performs steady and repetitive movements, taking shots according to the exposure settings predicted by various methods.}
\label{camera}
\vspace{-5pt}
\end{figure}

\noindent\textbf{Results for real capture.}
    We used a SONY Alpha 7C-\MakeUppercase{\romannumeral 2} to evaluate our model on real-world scenes. Subjects performed steady, repetitive movements while ISO and shutter speed were manually set for each capture, with the aperture fixed at f/2.8. The camera and subject positions remained unchanged during multi-exposure captures. As shown in \cref{camera}, our method provides noise control comparable to other baselines while effectively mitigating motion blur, which can impair the visual quality of the fused HDR image; for more results, see our supplementary materials.

\subsection{Ablation studies}

    \noindent\textbf{Effectiveness of penalty item.} We conduct ablation studies to validate the effectiveness of our reward design. All quantitative evaluations are conducted on the HDRV-Test dataset. We train our networks using only the construction reward and step penalty as the base model and test the effectiveness of $P_{\text{ghost}}$ and $P_{\text{priority}}$. As shown in \cref{ablation}, these two reward elements enhance our agent's performance by directing the model to focus on semantically important regions and moving areas. The former typically corresponds to the scene's focal point (\textit{e.g.} a person's face), while the latter involves regions at high risk of motion blur and ghosting artifacts. Qualitative evidence is provided in our supplementary materials. 

\begin{figure}[tbp]
\centering
\includegraphics[scale=0.34]{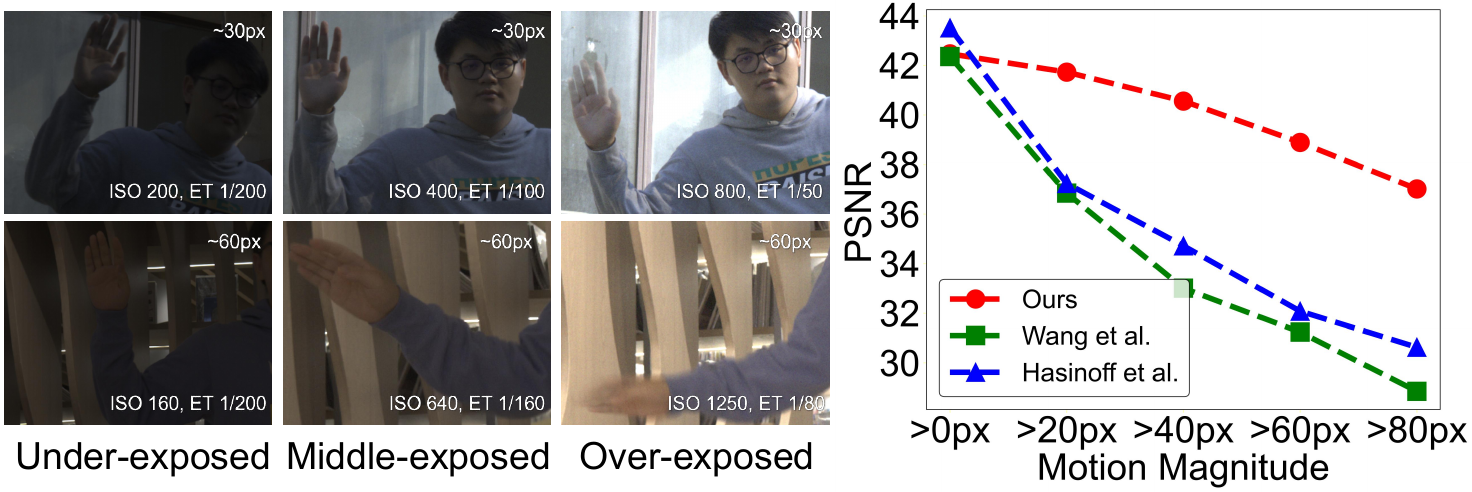}
\vspace{-18pt}
\caption{\textbf{Left:} Zoom-in details of the moving regions and our predicted ISO and shutter speed (seconds) for each LDR. Up: LDRs for scenes with an average motion level of around 30 pixels, Down: around 60 pixels. \textbf{Right:} Comparisons with baseline methods across different motion magnitude ranges. Tested on Real-HDRV~\cite{hdrv} dataset.}
\vspace{-18pt}
\label{motion}
\end{figure}

    \noindent\textbf{Robustness to dynamic scenes.} We offer a pair of examples from the Real-HDRV~\cite{hdrv} dataset with around 30 and 60 pixels of regional movement, respectively. As shown in \cref{motion}: \textbf{Left}, our trained agent tends to predict faster shutter speeds for scenes with stronger movement, validating the responsiveness of our agent in dynamic scenes. We assess our method's performance against previous auto-exposure methods \cite{hasinoff2010noise, wang2020learning} at varying motion magnitudes (\cref{motion}: \textbf{Right}). To create the evaluation dataset for robustness to different motion levels, we follow HDRFlow \cite{xu2024hdrflow}, using RAFT \cite{teed2020raft} to process dynamic scenes from HDRV and obtain optical flow maps. We then manually crop these images with reasonable flow predictions, dividing them into 128 $\times$ 128 blocks, and calculate the average motion magnitude for each block. Finally, we evaluate the PSNR of blocks corresponding to different motion magnitudes. As shown in \cref{motion}: \textbf{Right}, our AdaptiveAE demonstrates greater robustness than other methods as motion magnitude increases.

\section{Conclusion}
\label{sec:conclusion}
We introduce AdaptiveAE, which optimizes HDR exposure in dynamic settings using deep reinforcement learning, treating exposure bracketing as a Markov Decision Process. It autonomously adjusts ISO and shutter speed for a pretrained exposure fusion algorithm. Reward systems focus on moving and key regions, minimizing sequence lengths. Experiments show AdaptiveAE outperforms state-of-the-art methods in dynamic scenes while matching top models in static ones, allowing flexible HDR capture. Our analysis of noise models and exposure settings offers insights for future research, with plans to include adjustable apertures.

\noindent \textbf{Acknowledgement.} This work was supported by the National Key R\&D Program of China No.~2022ZD0160201, Shanghai Artificial Intelligence Laboratory, National Natural Science Foundation of China (Grant No.~62136001,~62088102), Beijing Natural Science Foundation (Grant No.~L233024), and Beijing Municipal Science \& Technology Commission, Administrative Commission of Zhongguancun Science Park (Grant No.~Z241100003524012). PKU-affiliated authors thank openbayes.com for providing computing resources.

\appendix
\renewcommand\thefigure{A\arabic{figure}}
\renewcommand\thetable{A\arabic{table}}  
\renewcommand\theequation{A\arabic{equation}}
\setcounter{section}{0}
\setcounter{equation}{0}
\setcounter{table}{0}
\setcounter{figure}{0}

\clearpage
\setcounter{page}{1}
\maketitlesupplementary

\begingroup 
\renewcommand{\thefootnote}{\fnsymbol{footnote}} 
\footnotetext[1]{This work was done during Tianyi Xu's internship at Shanghai AI Laboratory.} 
\footnotetext[2]{Corresponding authors.}
\endgroup

\section{More experimental results}
\subsection{More visual comparison results}

\begin{figure*}[tbp]
\centering
\includegraphics[scale=0.95]{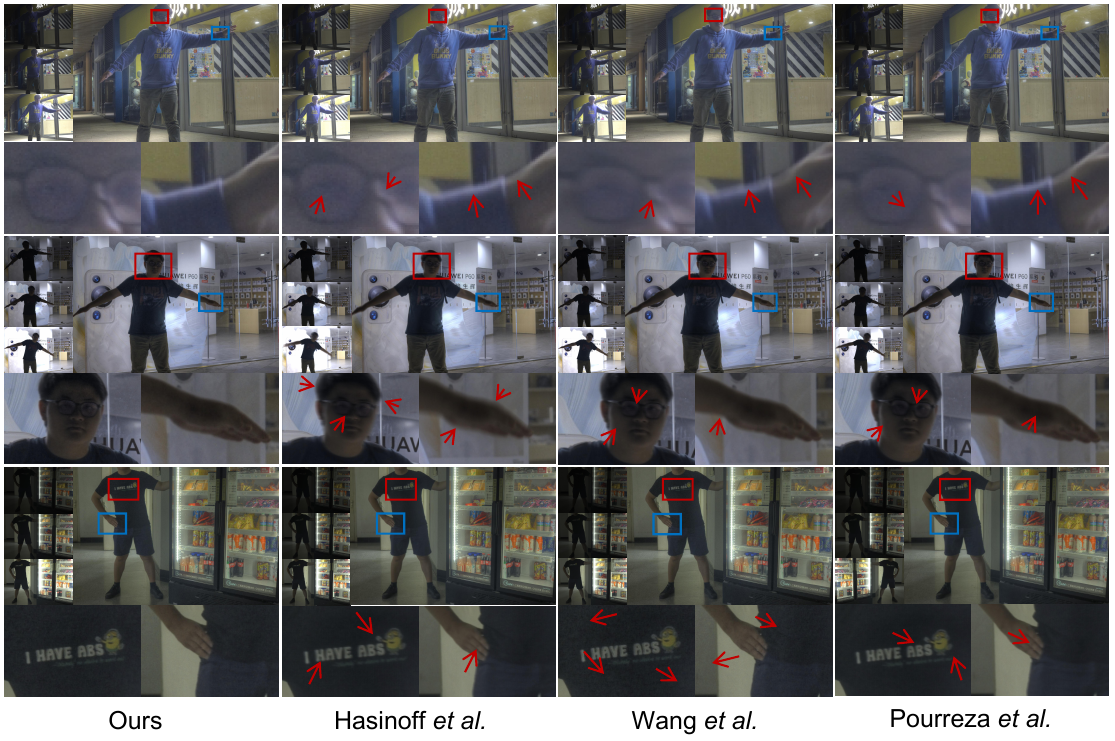}
\caption{Additional qualitative comparisons with other auto-exposure methods on HDRV dataset~\cite{hdrv}. Upper left: Predicted LDRs with varying ISO and shutter speed settings and synthesized using our image synthesis pipeline. Upper right: Fused HDR image using DeepHDR~\cite{huang2022real} and tone-mapped using Photomatix Enhancer. Below: Zoom-in results for tested methods.}
\label{sup_display}
\end{figure*}

As shown in \cref{sup_display}, we present additional qualitative results on the HDRV~\cite{hdrv} dataset. Our method strikes a balance between the impact of motion-related artifacts and the overall noise level, resulting in the best quality among all auto-exposure techniques. 

\begin{figure}[tbp]
\centering
\includegraphics[scale=0.3]{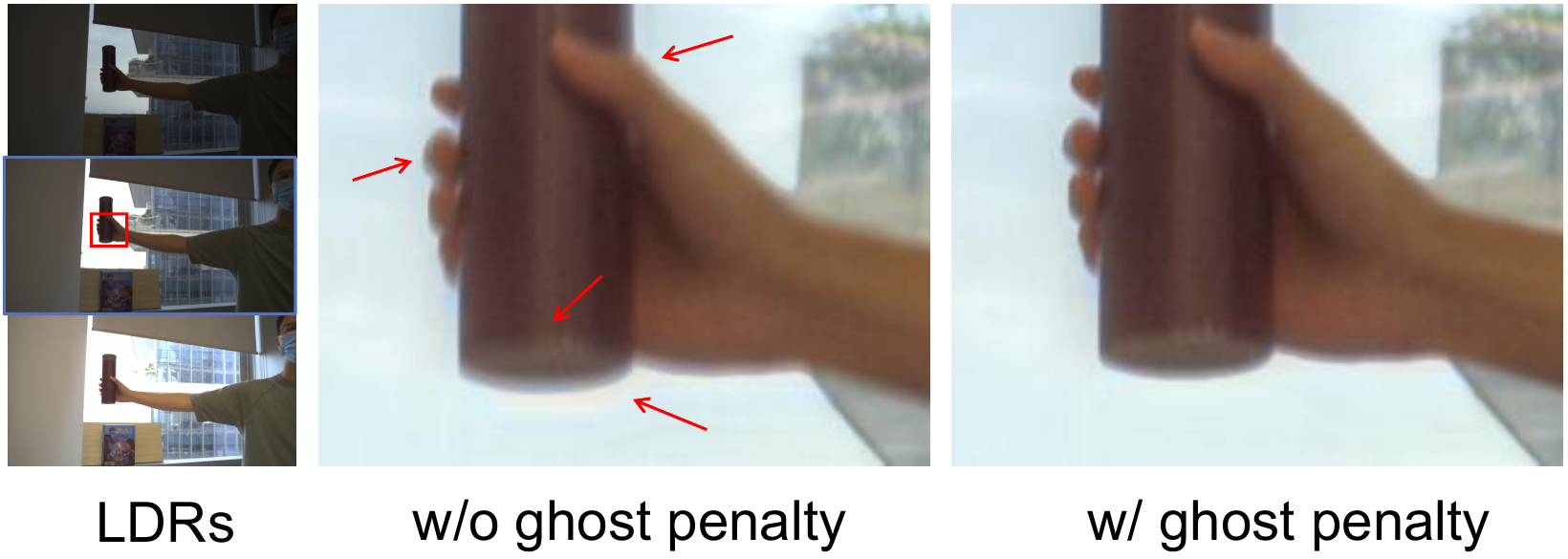}
\caption{Effectiveness of ghost penalty.}
\label{fig:ablation}
\end{figure}

\cref{fig:ablation} shows qualitative results for the ablation experiments on $P_{\text{ghost}}$ on a case in the DeepHDRVideo ~\cite{chen2021hdr} dataset; this penalty item effectively helps with mitigating ghosting and motion blur.

\begin{figure*}[tbp]
\centering
\includegraphics[scale=0.55]{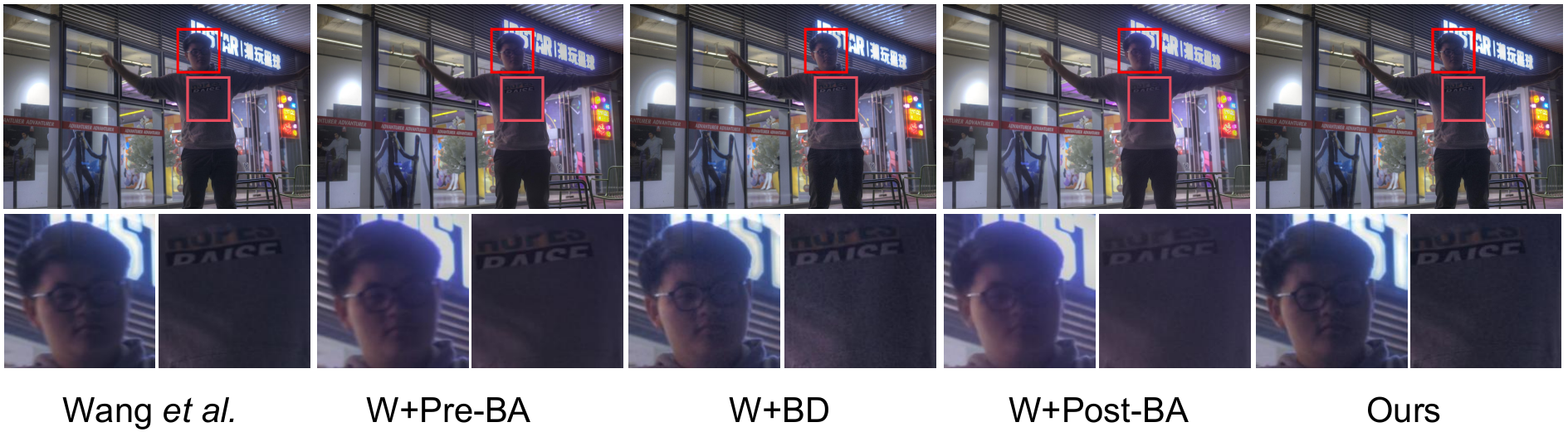}
\vspace{-4pt}
\caption{Necessity of considering motion blur during LDR capturing. We compared our method on the HDRV dataset~\cite{hdrv} with Wang~\etal~\cite{wang2020learning} combined with different post-processing deblurring methods. W denotes Wang~\etal, Pre-BA denotes applying BANet~\cite{tsai2022banet} to LDRs for fusion. BD denotes DeepHDR~\cite{huang2022real} trained on the HDRV-blur dataset. Post-BA denotes applying BANet to the tone-mapped fused HDR result.}
\label{fig:ablation2}
\end{figure*}

\subsection{Cross post-processing methods test} To validate the importance of addressing motion blur and ghosting during auto-exposure, we compared our results with Wang~\textit{et al.}~\cite{wang2020learning}, applying deblur methods at various stages: before, during, and after exposure fusion. For fairness, we trained the deblur models~\cite{barakat2008minimal, huang2022real} on our HDRV-blur dataset, created by adding random motion blur to HDR images from the HDRV dataset using our synthesis pipeline. For pre-fusion deblurring, BANet~\cite{tsai2022banet}, trained on HDRV-blur, was used to process the predicted LDRs before fusion with DeepHDR~\cite{huang2022real}. For fusion deblurring, we utilized DeepHDR's intrinsic deblurring ability, trained on HDRV-blur, without employing BANet. For post-fusion deblurring, BANet was applied after DeepHDR fusion. As shown in \cref{ablation_2} and \cref{fig:ablation2}, post-capture deblur minimally reduces blur in the fused HDR image but degrades static regions, highlighting the efficacy of addressing blur during LDR capture. 

\subsection{Analyzing the role of ISO}

\begin{figure}[tbp]
\centering
\includegraphics[scale=0.36]{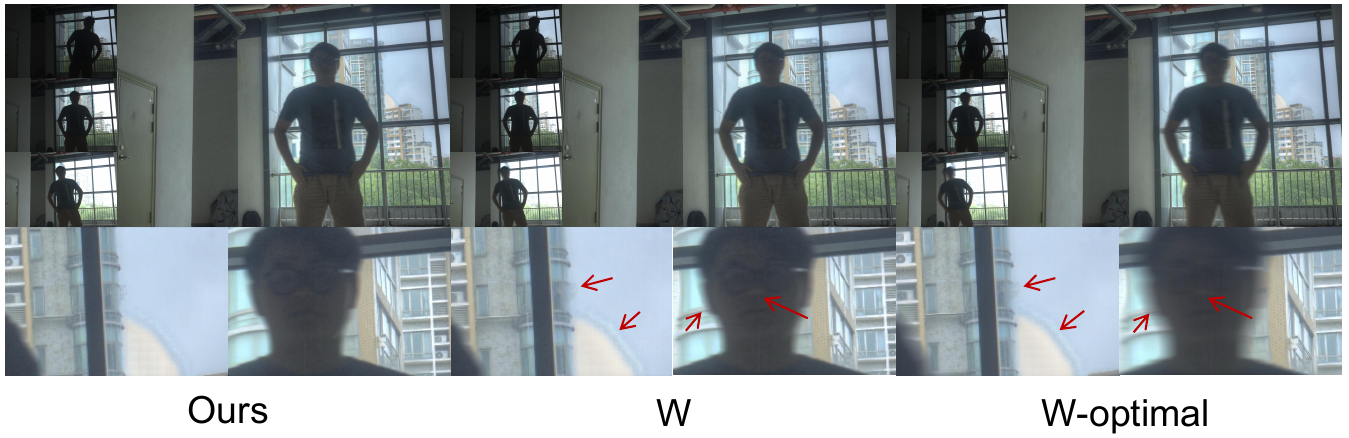}
\caption{Ablation on the role of ISO in fixed-ISO methods. We choose the PSNR-optimal fixed-ISO for Wang~\etal~\cite{wang2020learning} (represented by W) that optimizes the PSNR of generated HDR on HDRV-Test dataset~\cite{hdrv}, denoted as W-optimal. Upper left: Predicted LDRs with varying ISO and shutter speed settings and synthesized using our image synthesis pipeline. Upper right: Fused HDR image using DeepHDR~\cite{huang2022real} and tone-mapped using Photomatix Enhancer. Below: Zoom-in results for tested methods.}
\label{fig:iso_ablation}
\end{figure}

In our experiments (\cref{sec:experiments}), we set the ISO for fixed-ISO baselines to 200, as it serves as a standard choice in most scenarios. This raises the question of whether better results can be achieved by modifying the fixed ISO to an alternative value in the method proposed by Wang~\etal~\cite{wang2020learning}. To investigate this, we use Wang~\etal~\cite{wang2020learning} to predict the exposure values (EVs) for three low dynamic range images and systematically test all possible fixed-ISO settings to identify the value that maximizes the PSNR-$\mu$ on the test set. We denote this approach as W-optimal, where W refers to Wang~\etal~\cite{wang2020learning}. As illustrated in \cref{iso_ablation} and supported by the qualitative results in \cref{fig:iso_ablation}, utilizing the optimal fixed ISO results in slight performance improvement. However, this optimal ISO is highly dataset-specific and demonstrates very limited generalization capability, further validating the robustness and superiority of our proposed method over fixed-ISO approaches.

\subsection{More discussions on inference time}

Our RL agent executes in $<$5ms/scene on an NVIDIA RTX3080. The primary contributor to the latency, six LDR captures,
can mostly be eliminated if we use the frames cached in the preview buffer, also known as the ZSL (Zero Slag Latency) buffer, which is the de facto standard for mobile phones.
% can be further significantly accelerated by shifting capture time to the viewfinder buffer.
Utilizing an asynchronous camera driver, it has the potential to achieve real-time performance. In contrast, existing methods that use previously captured histograms for exposure prediction incur 30-70ms latency and are not robust to movement. Even without a viewfinder buffer, our inference speed can also be optimized with digital-overlap (DOL) sensors (to $<$100 ms/frame) and AE stats grid (around 32x24, downsampled from ISP 3A).

\begin{figure}[tbp]
\centering
\includegraphics[scale=0.265]{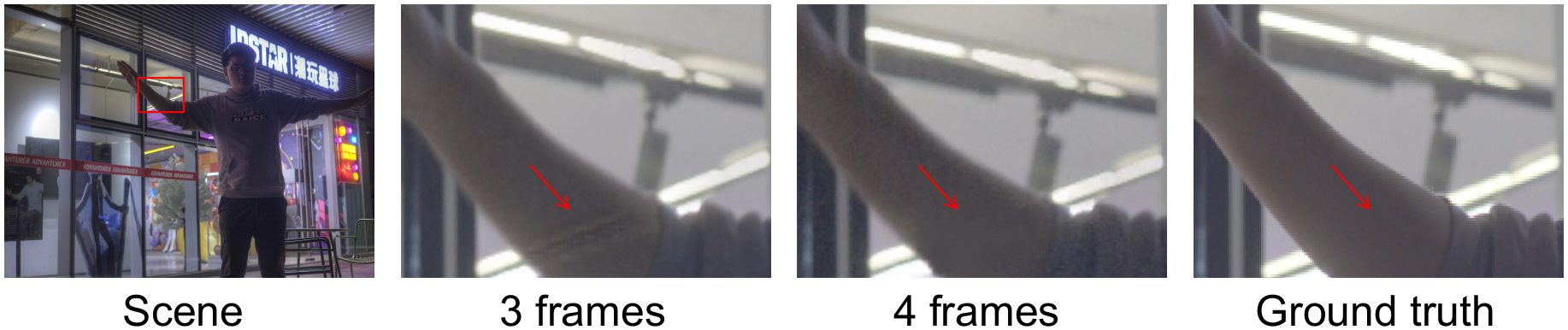}
\caption{An example scene for which our models give a 4-frame decision. 3-frame denotes a truncated version of the prediction, which has obvious ghosting patterns.}
\label{4step}
\end{figure}

\subsection{More frames}
Our design of the reward and the step penalty (\cref{step_penalty}) may result in a predicted exposure bracketing set containing more than three frames. \cref{4step} illustrates a case where our model makes a four-frame decision. Given our design of the step penalty, this occurs in only a small percentage of scenes with significant dynamic ranges and movements.

\begin{table}[tbp]
\centering
\small
\caption{Results for ablation studies for different deblur post-processing techniques. We use Wang~\etal~\cite{wang2020learning} as the base model (denoted as W), and Pre-BA denotes using BANet~\cite{tsai2022banet} to process the LDRs before exposure fusion. BD denotes using blur-aware DeepHDR~\cite{huang2022real} for exposure fusion, which is trained on the HDRV-blur dataset we synthesized from the HDRV~\cite{hdrv} dataset. Post-BA denotes using BANet to deblur the final tone-mapped HDR. \textbf{Bold}: The best.}
\setlength{\tabcolsep}{1pt} % Adjust this value as needed
\begin{tabular}{l|ccccc}
\hline
Model & PSNR-$\mu$ & SSIM-$\mu$ & PU-PSNR & PU-SSIM & \\ \hline
W & 36.46 & 0.8902 & 32.68 & 0.8933 \\
W+Pre-BA & 37.33 & 0.9095 & 33.24 & 0.9100 \\ 
W+BD & 37.01 & 0.9016 & 32.83 & 0.8972 \\
W+Post-BA & 37.25 & 0.9124 & 32.88 & 0.9023 & \\
Ours & \textbf{39.70} & \textbf{0.9408} & \textbf{34.67} & \textbf{0.9465} \\ \hline
\end{tabular}
\label{ablation_2}
\end{table}

\section{Details of the noise synthesis model}

We synthesize noise to the blurred HDR image $b_j^L$ according to our noise model, which is based on~\cite{hasinoff2010noise}. The quantity each pixel measures is the radiance level $\Phi$, in units of electrons per second. Therefore, the pixel value $I$ of a raw image can be expressed as:

\begin{equation}
   I = \text{min}\left\{\frac{\Phi T \times \text{ISO}}{U} + I_0 + n, I_{\text{max}}\right\},
\end{equation}
    where $T$ denotes the exposure time in seconds, $U$ is a camera-dependent constant, $I_0$ represents the electrons created by dark current, $n$ is the signal- and gain- dependent sensor noise and $I_{max}$ indicates the full well capacity.

    Conforming to the paradigm of \cite{hasinoff2010noise}, we model noise as a zero-mean variable, coming from three independent sources, including photon noise, which represents the Poisson distribution of photon arrivals and depends linearly upon the number of recorded electrons, $\Phi T$, readout noise, which comes from sensor readout, and analog-to-digital conversion(ADC) noise, which comes from the combined effect of amplifier and quantization. Hence, for pixels below the saturation level, we have:
    
\begin{equation}
    Var(n) = \frac{\Phi T \times \text{ISO}^2}{U^2} + \frac{\sigma_{\text{read}}^2 \times \text{ISO}^2}{U^2} + \sigma_{\text{ADC}}^2.
\label{noise_equation}
\end{equation}

    Note that the rationality of modeling ADC noise as independent of ISO lies in the fact that the quantization process, which could be represented by $q(x)$ in the following equation:
    
\begin{equation}
    q(x) = \text{min}(\left\lfloor {x+0.5} \right\rfloor, ADU),
\end{equation}
    where $ADU$ (Analog-to-Digital Units) denotes the maximum value that can be recorded by the sensor, for a target camera that records scenes as $b$-bits raw images, $ADU = 2^b - 1$, this $q$ function is independent of ISO settings. The post-amplifier noise is also naturally independent of the foreground imaging settings. 

    Following our noise model, we can synthesize the corresponding noise with the decided ISO and shutter speed to the blurred HDR $b_j^L$, thereby obtaining the LDR image $l_j^\mathcal{T}$. This noise model facilitates the synthesis of LDR images with various ISO and shutter speed settings. Moreover, it accurately simulates the actual noise that arises in photography, helping our model to exhibit good generalization abilities on various datasets and real data. 

\begin{table}[tbp]
\centering
\small
\caption{Ablation study of the impact of ISO on fixed-ISO methods on HDRV~\cite{hdrv} dataset. $W$ denotes Wang~\etal~\cite{wang2020learning} and $W$-optimal denotes setting the fixed ISO of Wang~\etal~\cite{wang2020learning} to optimal value for SNR cross all available ISOs. \textbf{Bold}:best.}
\setlength{\tabcolsep}{1pt} % Adjust this value as needed
\begin{tabular}{l|ccccc}
\hline
Model & PSNR-$\mu$ & SSIM-$\mu$ & PU-PSNR & PU-SSIM & \\ \hline
$W$ & 36.46 & 0.8902 & 32.68 & 0.8933 \\
$W$-optimal & 37.64 & 0.9033 & 33.01 & 0.9058 \\
Ours & \textbf{39.70} & \textbf{0.9208} & \textbf{34.67} & \textbf{0.9465} \\ \hline
\end{tabular}
\label{iso_ablation}
\end{table}
    
    Denoting the entire image synthesis process, which consists of motion blur synthesis and adding noise, as $\mathcal{S}$, and the corresponding LDR output as $l_j^\mathcal{T}$, we have:

\begin{equation}
    l_j^\mathcal{T} = \mathcal{S}(f_i^\mathcal{T}, f_{i+1}^\mathcal{T}, (\text{ISO}_j, T_j)).
\end{equation}
    where $\text{ISO}_j$ and $T_j$ are bracketed to denote that they are a pair of camera settings.

\section{Network details}
The architecture of our proposed AdaptiveAE network comprises two primary components: a Policy Network and a Value Network. The Policy Network is responsible for producing two output layers: one with 24 units for ISO selection and another with 19 units for shutter speed selection. Specifically, the ISO space consists of 24 possible settings—\{50, 64, 80, 100, 125, 160, 200, 250, 320, 400, 500, 640, 800, 1000, 1250, 1600, 2000, 2500, 3200, 4000, 5000, 6400, 8000, 10000\}—and the shutter speed space contains 19 possible values—\{1/30, 1/40, 1/50, 1/60, 1/80, 1/100, 1/125, 1/160, 1/200, 1/250, 1/320, 1/400, 1/500, 1/640, 1/800, 1/1000, 1/1250, 1/1600, 1/2000\}. The Policy Network employs softmax activation functions for both outputs, providing probability distributions over possible ISO and shutter speed configurations.
In contrast, the Value Network outputs a single-unit layer, which estimates the state value. To ensure non-negative outputs, the Value Network incorporates a ReLU activation function. The separation of the Policy and Value Networks facilitates efficient decision-making by modeling both action distribution and state evaluation independently, allowing the system to adapt effectively to varying exposure conditions.

\subsection{Semantic feature branch}
The semantic feature branch leverages pre-trained AlexNet features, initially with a dimensionality of 4096. We apply this branch to the median-exposed LDR image from the input set. These semantic representations are transformed using a two-layer fully connected architecture. The first layer comprises 1024 neurons, while the second has 256 neurons, both with ReLU activation.

\subsection{Irradiance feature branch}
The irradiance feature branch processes exposure information from multiple LDR images by extracting histograms from each LDR image separately and concatenating them along the channel dimension. This multi-exposure histogram data is processed through three sequential 1D convolutional layers: the first with 128 filters, the second with 256 filters, and the third with 512 filters, all using a kernel size of 4 and a stride of 4. Following this, two fully connected layers with 1024 and 256 neurons, respectively, process the features, maintaining ReLU activation throughout.

\subsection{Stage encoding branch}
The stage encoding branch introduces a temporal dimension to the network by encoding both the current exposure iteration and the total planned exposures. It processes a two-dimensional input (current stage, total stages) through two layers: the first with 32 neurons and the second with 64 neurons, both activated by ReLU functions. This enhancement allows the network to adapt its strategy based on the remaining exposure budget.

\subsection{Feature fusion mechanism}
Features from the multiple LDR inputs, semantic, irradiance, and stage encoding branches are concatenated and processed through two fusion layers for comprehensive integration. The first fusion layer includes 512 neurons, followed by a second layer with 256 neurons, both using ReLU activation. This thorough fusion of features equips the network with the capacity to synthesize multi-modal information, thereby enhancing predictive accuracy.

Despite accepting multiple LDR inputs directly into each processing branch, the architecture maintains computational efficiency with approximately 7-8 million parameters, achieving inference times under 10 milliseconds, making it suitable for real-time applications in computational photography and image signal processing.

\end{document}